\documentclass[sigconf]{acmart}

\usepackage{multirow}
\usepackage{stfloats}

\AtBeginDocument{%
  \providecommand\BibTeX{{%
    \normalfont B\kern-0.5em{\scshape i\kern-0.25em b}\kern-0.8em\TeX}}}

\copyrightyear{2020} 
\acmYear{2020} 
\setcopyright{acmcopyright}
\acmConference[CIKM '20]{Proceedings of the 29th ACM International Conference on Information and Knowledge Management}{October 19--23, 2020}{Virtual Event, Ireland}
\acmBooktitle{Proceedings of the 29th ACM International Conference on Information and Knowledge Management (CIKM '20), October 19--23, 2020, Virtual Event, Ireland}
\acmPrice{15.00}
\acmDOI{10.1145/3340531.3412688}
\acmISBN{978-1-4503-6859-9/20/10}

\begin{document}
\fancyhead{}
\title{AutoADR: Automatic Model Design for Ad Relevance}

\author{Yiren Chen}
\authornote{Authors contributed equally to this work.}
\authornote{This work was done when the author was an intern at Microsoft Research Asia.}
\authornote{Authors are affiliated with Key Laboratory of Machine Perception, MOE, School of EECS, Peking University.}
\affiliation{\institution{Peking University}}
\email{yrchen92@pku.edu.cn}
\author{Yaming Yang}
\authornotemark[1]
\affiliation{\institution{Microsoft Research Asia}}
\email{yayaming@microsoft.com}
\author{Hong Sun}
\authornotemark[1]
\affiliation{\institution{Microsoft}}
\email{hosu@microsoft.com}

\author{Yujing Wang}
\authornotemark[3]
\affiliation{\institution{Microsoft Research Asia}}
\email{yujwang@microsoft.com}

\author{Yu Xu, Wei Shen}
\affiliation{\institution{Microsoft}}
\email{{xuyu, sashen}@microsoft.com}

\author{Rong Zhou}
\affiliation{\institution{Microsoft Research Asia}}
\email{v-roz@microsoft.com}

\author{Yunhai Tong}
\authornotemark[3]
\affiliation{\institution{Peking University}}
\email{yhtong@pku.edu.cn}

\author{Jing Bai}
\affiliation{\institution{Microsoft}}
\email{jbai@microsoft.com}

\author{Ruofei Zhang}
\affiliation{\institution{Microsoft}}
\email{bzhang@microsoft.com}


\renewcommand{\shortauthors}{Yiren Chen, Yaming Yang and Hong Sun, et al.}

\begin{abstract}
Large-scale pre-trained models have attracted extensive attention in the research community and shown promising results on various tasks of natural language processing. However, these pre-trained models are memory and computation intensive, hindering their deployment into industrial online systems like Ad Relevance. Meanwhile, how to design an effective yet efficient model architecture is another challenging problem in online Ad Relevance. Recently, AutoML shed new lights on architecture design, but how to integrate it with pre-trained language models remains unsettled. In this paper, we propose AutoADR (Automatic model design for AD Relevance) --- a novel end-to-end framework to address this challenge, and share our experience to ship these cutting-edge techniques into online Ad Relevance system at Microsoft Bing.
Specifically, AutoADR leverages a one-shot neural architecture search algorithm to find a tailored network architecture for Ad Relevance. The search process is simultaneously guided by knowledge distillation from a large pre-trained teacher model (e.g. BERT), while taking the online serving constraints (e.g. memory and latency) into consideration. 
We add the model designed by AutoADR as a sub-model into the production Ad Relevance model. This additional sub-model improves the Precision-Recall AUC (PR AUC) on top of the original Ad Relevance model by \textbf{2.65X} of the normalized shipping bar. More importantly, adding this automatically designed sub-model leads to a statistically significant \textbf{4.6\%} Bad-Ad ratio reduction in online A/B testing. 
This model has been shipped into Microsoft Bing Ad Relevance Production model.
\end{abstract}




\maketitle

\section{Introduction}
Large-scale pre-trained models, like BERT, have demonstrated their superiority in many Natural Language Processing (NLP) tasks, such as text classification~\cite{chang2019x}, reading comprehension~\cite{zhang2020retrospective} and machine translation \cite{zhu2020incorporating}. In the meanwhile, AutoML has attracted extensive attention in the research community and shown promising results on various academic datasets in computer vision and natural language processing. 
For instance, NasNet~\cite{zoph2018learning} proposed a search space to find transferable architectures for scalable image recognition. EfficientNet \cite{tan2019efficientnet} proposed a simple and highly effective compound scaling method that enables easy scaling up of a backbone model and beats the performances of human-designed image recognition architectures by a large margin. In the NLP domain, some of us proposed TextNAS~\cite{wang2019textnas}, which designed a novel search space tailored for text representation and achieved state-of-the-art performances on various natural language processing tasks when pre-training was not applied.

Motivated by recent research progresses, we are aiming to ship these cutting-edge techniques into our online Ad Relevance model in Microsoft Bing. Ad Relevance measures how close an Ad is to the user's search query. It is crucial to the ecosystem of online advertising, as it affects the search engine's revenue, user experience, and advertiser satisfaction directly.
Despite the effectiveness of existing approaches, we still face two major challenges to apply them successfully to our online Ad Relevance system. 
\begin{itemize}
    \item Firstly, the pre-trained models are memory and computation intensive, hindering their deployment into industrial online systems directly. We need to generate a tailored model for each application that fulfills the memory and latency constraint while retaining the original accuracy of pre-trained models to the largest extent.
    \item Secondly, although AutoML and pre-training themselves have demonstrated good performances separately in the literature, few existing works have explored a joint solution.
    \textit{How to unify the power of AutoML and pre-training models collaboratively to improve the model performance?}
\end{itemize}
In this paper, we propose the AutoADR (Automatic model design for AD Relevance) framework to address the above questions and challenges. The framework is based on knowledge distillation~\cite{hinton2015distilling}.
In the AutoADR framework, we can first improve the performance of the pre-trained teacher model with large parameter size and model ensemble without considering the memory and latency constraints. Next, we conduct neural architecture search to find the best architecture that balances accuracy and efficiency. The NAS procedure is performed jointly with knowledge distillation in an iterative manner, so that NAS and knowledge distillation will collaborate with each other to find a tailored model architecture to achieve our goals on both accuracy and efficiency.
Moreover, following TextNAS~\cite{wang2019textnas}, we leverage a customized search space for text representation, which consists of a multi-path mixture of convolutional, recurrent, pooling, and self-attention layers. In this way, we can explore the best composition of different layers for the Ad Relevance prediction problem. 

We conduct offline experiments with a real-world dataset for Ad Relevance collected from Microsoft Bing search log. As shown in the results, the model designed and trained by the AutoADR pipeline demonstrates superior performances compared to the baseline models. We also apply it to the production model with evaluation results showing the effectiveness of this model. Specifically, AutoADR leads to a normalized 2.65\% Precision-Recall AUC lift when integrated with the production model, which is beyond the normalized 1\% shipping bar. More importantly, it achieves a statistically significant 4.6\% reduction of Bad-Ad ratio in online A/B testing. 
This new model has been shipped into Microsoft Bing Ad Relevance Production model.

The \textbf{main contributions} of this paper are summarized below. 
\begin{enumerate}
    \item We propose AutoADR, a general end-to-end framework for automatic model design with knowledge distillation. It takes the advantages of both AutoML and knowledge distillation to build a customized model for a specific task with certain constraints.
    \item We apply AutoADR to Ad Relevance scenario and demonstrate its effectiveness for designing a tailored sub-model for Ad Relevance. Compared to other human-crafted architectures, the model designed by AutoADR shows better performances and trade-offs on both accuracy and latency. 
    \item After adding the derived sub-model to the production pipeline, the Bad-Ad ratio has been reduced significantly during online A/B testing. To the best of our knowledge, this is one of the first works that demonstrate the success of AutoML in a large-scale online application.
\end{enumerate}

The rest of this paper is organized as follows. Related works are reviewed in the next section. Section 3 describes the context of Ad Relevance in online advertising. In Section 4, we provide the methodology of the AutoADR framework in detail. Section 5 discusses the experimental results and introduces the deployment of AutoADR in the production system. The last section gives the conclusion and future work.

\section{Related Work}

\textbf{Pre-trained Language Models} has achieved significant improvement in a wide range of NLP tasks by learning deep representations from a large-scale corpus~\cite{chang2019x, zhu2020incorporating, yang2019end}. However, they usually have large parameter sizes and complex model structures, which hinders their deployment in real-time applications due to memory and latency constraints. Therefore, a variety of works aim to compress BERT into faster and lighter ones. Motivated by knowledge distillation, PKD-BERT~\cite{sun2019patient} and DistilBERT~\cite{sanh2019distilbert} compress BERT into shallow structures by distilling information during fine-tuning and pre-training phase respectively. TinyBERT~\cite{jiao2019tinybert} utilizes the two-stage knowledge distillation, transferring embedding, and hidden attention information from a teacher model to a student model. Apart from Transformer, BiLSTM~\cite{tang2019natural} and CNN~\cite{chen2020adabert} are also considered as lighter alternatives to build deep light networks for specific NLP tasks. In our work, we propose an AutoADR framework to search structures suitable for distilling knowledge of BERT.

\textbf{Neural Architecture Search} encodes the network structure into numerical sequences and searches for a sequence corresponding to an optimal architecture in an automated way with as little human intervention as possible~\cite{zoph2016neural}.
Although it achieves competitive performance on a specific task, the search process usually requires huge computation resources (thousands of GPU hours). To make NAS more efficient, weight sharing strategies~\cite{liu2018darts, pham2018efficient, bender2018understanding, guo2019single} are applied to speed up the search and evaluation stages. A supernet consisting of all possible architectures in a given search space is encoded. All structures share the weights in the same nodes. Once the supernet is trained, each sampled structure can be directly and quickly evaluated without training from scratch. In our work, we also leverage the weight sharing strategy and build a one-shot model~\cite{guo2019single} with TextNAS~\cite{ wang2019textnas} search space, while random search is adopted to sample architectures. Besides, we incorporate knowledge distillation and efficiency constraints as search hints to obtain effective and light student models.

\textbf{Ad Relevance} measures how relevant advertiser-sponsored ads are to user-issued queries. Different from other relevance tasks, the text length of the queries given by users is usually short in the sponsored search engine, making it difficult to match the user’s intent. To address the problem, some works focus on query expansion and rewriting to enrich query information. 
\cite{gao2012learning} constructs a simple query expansion model that incorporates the lexicon model. \cite{bai2018scalable} proposes a novel embedding of queries to improve ad matching in Sponsored Search, which is generated from constituent word n-gram embeddings. On the other hand, text representation methods are also explored by researchers to improve the performance of this problem. C-DSSM~\cite{shen2014latent} is a well-known learning-to-match paradigm, which leverages a convolutional neural architecture to capture the query intent. Jointly modeling query content as well as its context, \cite{sordoni2015hierarchical} designs a novel hierarchical recurrent encoder-decoder architecture, which is sensitive to the order of queries. 
Different from existing works, our architecture is obtained from the architecture search process, which is more complicated but performs well in both online and offline metrics.

\section{Application to Ad Relevance}
\subsection{System Overview}
\begin{figure}[t]
    \centering
    \includegraphics[width=\linewidth]{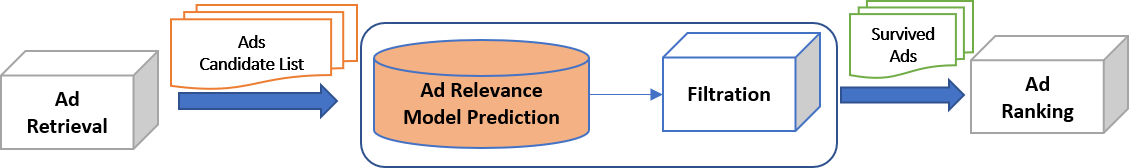}
    \caption{Simplified framework of online ads serving pipeline}
    \label{fig:ad_serving}
    \vspace{-0.5cm}
\end{figure}
Online advertising serves as a bridge connecting the user's search intents with ads provided by advertisers. The overall system is complicated and Figure \ref{fig:ad_serving} shows a simplified framework. In general, it contains three key components: 

\textbf{Ad Retrieval component} performs the initial retrieval step with techniques like Information Retrieval (IR) to generate a large candidate list for a given user query. It favors recall over precision to ensure all potentially related ads can be retrieved from ad corpus.

\textbf{Ad Relevance component} measures relevance between query and ads, and ensures that ads passing to downstream ranking component are relevant to user query. As Figure~\ref{fig:ad_serving} shows, Ad Relevance component performs filtration on irrelevant ads, and a better Ad Relevance model will yield to a higher precision in such filtration decisions, which leads to lower Bad-Ad ratio at final ad impressions.

\textbf{Ad Ranking component} makes final decisions on what ads are shown and the order of showing them. It makes a comprehensive decision based on multiple signals, including user intent, the probability of the user’s click-through rate (CTR), Ad Relevance score, bidding price, etc.

Among all the three components, Ad Relevance is indispensable as it plays a crucial role to improve relevancy between ads and user queries. Showing irrelevant ads will lead to bad user experience and poor advertiser satisfaction. To improve Ad Relevance metrics, it heavily relies on Natural Language Processing (NLP) techniques to understand the intent of user queries and ads content. In this work, we will use Ad Relevance as our main task to demonstrate the effectiveness of our proposed method. 
However, our method is not limited to Ad Relevance and can be extended to other relevance tasks. 
In the future, we plan to expand its application to other components in online search and advertising systems.

\subsection{Ad Relevance Model}
\label{ad_relevance_production}
\begin{figure}
    \centering
    \includegraphics[width=6cm]{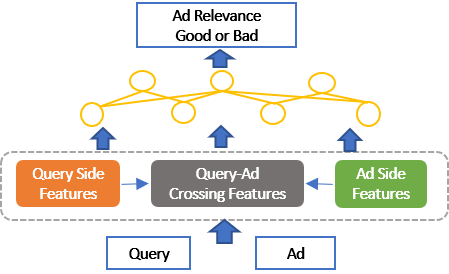}
    \caption{Overall architecture of Ad Relevance modeling}
    \label{fig:ad_relevance_arc}
    \vspace{-0.5cm}
\end{figure}

Ad Relevance model measures how much a given query-ad pair is semantically-related to each other, and it aims to filter as many unrelated ads as possible at online serving. The overall architecture of an Ad Relevance model is illustrated in Figure \ref{fig:ad_relevance_arc}. In this framework, the model is trained based on human labeled query-ad pairs with relevance labels being good or bad. The model itself is a multi-layer neural network. This model ensembles features mined from query and ad, as well as crossing features between those two. Those features can be either manually designed ones or representations learned from sub-models. In this work, we will study the impact of adding model learned by our proposed method as a new sub-model into production Ad Relevance model. 

When building the sub-model used in Ad Relevance, we make some specialized design on model structure. Given a query, there could be thousands of ads whose relevance need to be measured. For high-complexity models like BERT, utilizing a single encoder to take one query-ad pair as input each time will lead to unbearable serving cost. 
Therefore we leverage a structure similar to \cite{shen2014latent}, where the query and ad are encoded separately and then fed into a crossing layer to compute their interactions. 
Such design makes it possible to decouple the processing of query and ad, specifically, the ad-side vectors can be pre-calculated offline, while the query-side vectors are calculated online and some head queries can be cached in advance. 
After getting both vectors, a crossing model calculates the final score of AutoADR sub-model, which is sent to Ad Relevance model as an input. In this way, we can save tremendous online computation costs while still benefit from representations learned by the encoder model. 

\section{AutoADR}

\subsection{Overview}

\begin{figure}[t]
    \centering
    \includegraphics[width=\linewidth]{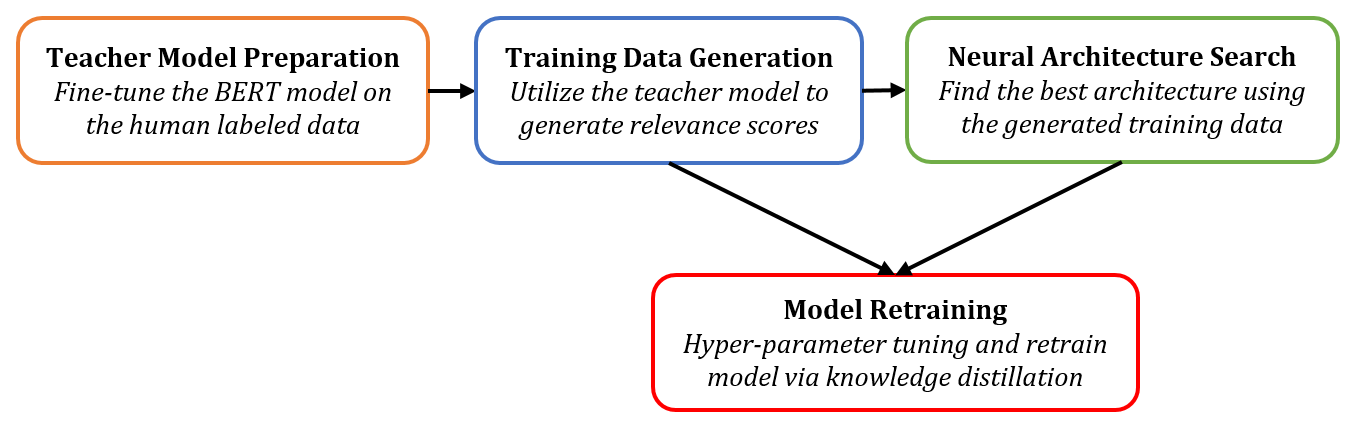}
    \caption{The overall processing pipeline of AutoADR}
    \label{fig:process_pipeline}
    \vspace{-0.5cm}
\end{figure}

In this section, we present the AutoADR pipeline, which aims to automatically design an effective and efficient sub-model for Ad Relevance while taking advantage of the pre-trained model as much as possible. This pipeline leverages one-shot neural architecture search to find the optimal architecture and incorporates knowledge distillation with efficiency constraints. The overview of the AutoADR pipeline is shown in Figure \ref{fig:process_pipeline}, which consists of four procedures.

\textbf{Teacher Model Preparation}: To apply knowledge distillation, we first need to train a teacher model. We choose BERT-large model as the teacher since it has achieved outstanding performances on many NLP tasks. 
Deploying such a big model in the production system is challenging due to resource limitation and latency constraints. 
However, it can promote the performance of a student model by providing soft predictions as guide and in this way we can transfer its power into production.

\textbf{Training Data Generation}: We mine hundreds of millions of impressed query-ad pairs from the search engine log as training data. 
We use scores of those data predicted by the teacher model as soft targets to guide the searching and retraining process.

\textbf{Neural Architecture Search}: The task is to find an effective and efficient architecture for AutoADR sub-model through teacher-student framework.
We adopt the search space in TextNAS~\cite{wang2019textnas} and build a corresponding network subsuming all possible architectures, which is called supernet~\cite{guo2019single} for the one-shot search algorithm. To apply knowledge distillation, we train the supernet with uniform sampling under the guidance of the soft predictions from teacher model. 
In the architecture searching phase, the random search algorithm is employed to select the best architecture from thousands of sampled architectures.
It could be replaced by other methods, such as reinforcement learning and evolutionary algorithms. 

\textbf{Model Retraining}: After obtaining the optimal architecture in the search process, a hyper-parameter search procedure is conducted on the validation set to seek for the best configuration. Finally, we retrain the model with the specific configuration and full training data in the knowledge distillation framework. We adopt the Tree-structured Parzen Estimator Approach (TPE)~\cite{Bergstra2011Algorithms}, which has faster speed and better performance compared to other Bayesian optimization algorithms in the high dimension search space. The valid search spaces for hidden dimensions are forced to fulfill the memory and latency constraints. 

\subsection{Model Architecture}

\begin{figure}[t]
    \centering
    \includegraphics[width=\linewidth]{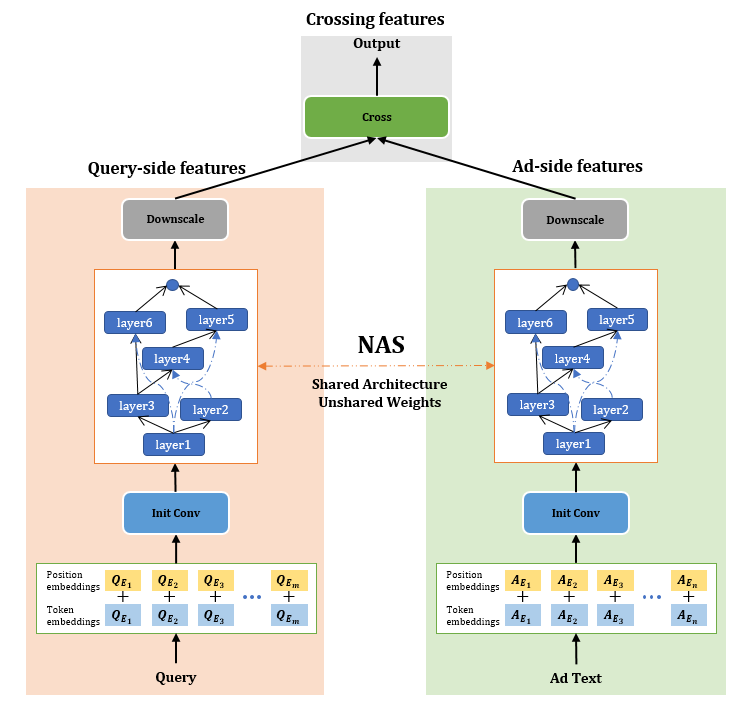}
    \caption{Overall structure of AutoADR model} 
    \label{fig:model_architecture}
    \vspace{-0.5cm}
\end{figure}

We leverage AutoADR to design a sub-model to improve the performance of production Ad Relevance model. Here we present the overall model architecture to be searched by AutoADR. Specifically, we leverage a twin-tower architecture as required in our production system for efficiency consideration~(Section \ref{ad_relevance_production}). 
As shown in Figure \ref{fig:model_architecture}, the architecture consists of two multi-layer encoders to be designed by neural architecture search. Their architectures are shared, while the weights are distinct and separately learned. Based on the query and Ad representation outputs from two encoding modules, we utilize a crossing layer to capture their interactions and produce the final prediction score. Each component of the model will be discussed in the sub-sections below.

\subsubsection{\textbf{Embedding Layer}}

Two input sentences, query and ad content, are encoded separately and fed into corresponding encoders. 
We utilize the tri-letter based word embedding introduced in~\cite{shen2014learning} as the token embeddings. Specifically, a word is firstly segmented into a sequence of tri-letter tokens after adding word boundary symbols(\#). Then, we sum over all word tri-letter features uniformly to get the word embedding. There are about 50K tokens in the vocabulary, which can constitute most words appeared in the online advertising system. 
Compared to WordPiece used in vanilla BERT~\cite{devlin2019bert}, tri-letter is more efficient to segment words without the recursive process, which is beneficial to industrial online systems where latency is a critical constraint. Learnable position embeddings are also included in the model. For a given word, its input representation is constructed by summing the corresponding token embeddings and position embeddings.

\subsubsection{\textbf{Initial Convolutional Layer}}

A 1-D convolutional layer is stacked upon the embedding layer, thus decoupling the hidden size from the embedding size. Specific configurations of the convolutional layer are: the kernel size is 1, the number of input channels equals to the embedding size while the number of the output channels is equal to the hidden size. With this separation, the hidden size could be increased easily to enlarge the model capacity without significantly increasing the parameter size of the vocabulary embeddings.

\subsubsection{\textbf{Encoding Layer}}

As shown in Figure \ref{fig:model_architecture}, the encoding layer consists of two multi-layer encoders that produce sentence representations for query and ad content respectively. It is noteworthy that the architectures of two encoders are shared during the search and evaluation processes, while their parameters are learned separately to capture distinct information. The architecture of encoding module is designed by neural architecture search with knowledge distillation, which will be introduced in Section \ref{NAS_KD}. 
The encoding module has $k$ layers, incorporating four common categories of candidate operations: convolutional layers, recurrent layers, pooling layers, and multi-head self-attention layers. The shape of input in each layer keeps the same, so that every module can be stacked freely with skip connections. In our experiments, we set $k = 6$ as a trade-off between expressiveness and efficiency.
Following TextNAS~\cite{wang2019textnas}, the network architecture of the encoder supports multi-path ensemble, which is a common design principle of manual networks.

\subsubsection{\textbf{Downscale Layer}}

A downscale layer is added to separate the representation dimension from the hidden size. Specifically, the downscale layer linearly projects the query representation and ad representation to a compact space. Such design is necessary since the representation vectors of query and ad will be fed to an online processing module for further combination, and the representation should be compact enough to satisfy the memory and latency constraints.

\subsubsection{\textbf{Crossing Layer}}

To enhance the relationship between query-ad pairs, we concatenate the embeddings of two sentences and also add their absolute difference and element-wise product~\cite{mou2016natural} as the input of the multi-layer perceptron (MLP) classifier:
\begin{equation}
x = Concat(q, a, |q - a|, q \odot a)
\end{equation}

\noindent where $q$ and $a$ represent the embedding of query and ad content respectively, $\odot$ is the element-wise product. $Concat$ is the concatenation operation. Then, $x$ will be fed into the MLP classifier with two hidden layers and ReLU activation. The shortcut connection is adopted to overcome over-fitting and gradient vanishing problems. At last, we generate the output through Sigmoid activation.

\subsection{NAS with Knowledge Distillation}
\label{NAS_KD}

We propose a joint solution of knowledge distillation and neural architecture search for Ad Relevance. Neural architecture search looks for better architecture for query and ad encoders, while knowledge distillation exploits useful knowledge from pre-trained models to facilitate representation learning. These two procedures are performed simultaneously and collaboratively towards a better performance. 

\begin{figure}[t]
    \centering
    \includegraphics[scale=0.165]{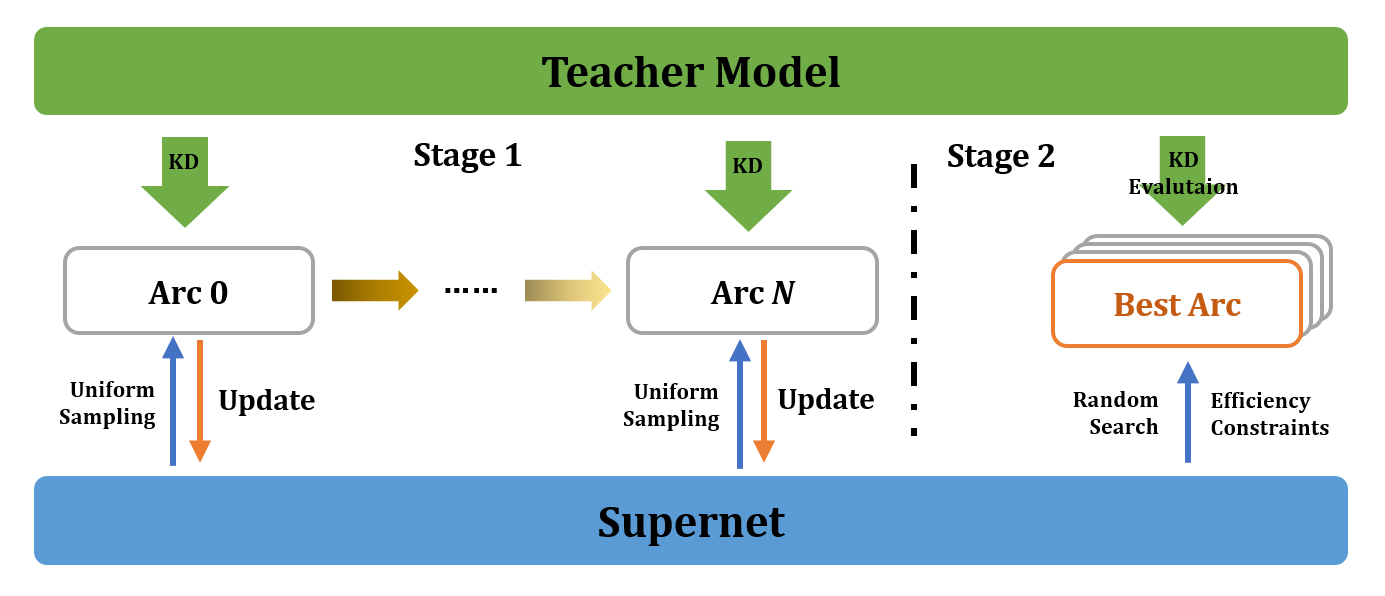}
    \caption{One-shot architecture search with knowledge distillation}
    \label{fig:search_algorithm}
    \vspace{-0.5cm}
\end{figure}

As illustrated in Figure \ref{fig:search_algorithm}, we present a two-stage searching framework for selecting a target architecture with knowledge distillation. In the first stage, we pick a few architectures from the supernet~\cite{guo2019single} defined by the search space. Then, they are trained one by one under the guidance of the teacher model through a knowledge distillation procedure iteratively. In the second stage, once the performance of supernet converges, we perform a final search to find out the best architecture from the candidate list. Latency constraints are also considered in the final search step. We will discuss the details of the process in the rest of this section.

\subsubsection{\textbf{Search Space}}

The architecture of neural networks can be depicted by a general directed acyclic graph (DAG). Each layer in the model is taken as a node, and the connection between layers is presented as the edge. We define the search space as a supernet $\mathcal{A}$, where a diverse set of candidate architectures can be captured by sampling a certain number of nodes and edges.
Following TextNAS~\cite{wang2019textnas}, we build the search space by incorporating multi-path ensembles and a mixture of different operations, including \textbf{convolution}, \textbf{pooling}, \textbf{recurrent} and \textbf{self-attention}. In this way, we can resort to the neural architecture search algorithm to find a tailored solution for Ad Relevance, which could benefit jointly from different kinds of layers. 

Specifically, we choose 1-D standard convolution with kernel size  $\{1, 3, 5, 7\}$ and apply ReLU-Conv-BatchNorm structure once it has been added. Maximum and average pooling are included and their filter size is set as 3. We use the “SAME” padding to guarantee that the number of output filters is equal to the input dimension. We select bi-directional GRU layer as our \emph{recurrent} cell implementation, which sums the output vectors of two opposite directions. \emph{Self-attention} cell is defined as multi-head self-attention layer, which is a major component in the neural network of Transformer \cite{vaswani2017attention}. The number of attention heads is set as 8 in all experiments. Each layer has the same shape, so that one can stack multiple layers freely with skip connections.

\subsubsection{\textbf{Knowledge Distillation}}

Knowledge distillation is a compression technique in which a compact student model can be trained to mimic the behavior of a large teacher model. In AutoADR, the pre-trained model is fine-tuned on the human labeled relevance data and is then served as teacher model to score a collection of impressed query-ad pairs. Through knowledge distillation, We aim to obtain an optimal architecture $\alpha \in \mathcal{A} $ by minimizing the following cross-entropy loss function:
\begin{align}
    & \mathcal{L}_{KD}=-\sum\limits_{i=1}^N(y_i\log(p_i)+(1-y_i)\log(1-p_i)) \\
    & y_i=\frac{\exp(z_i/T)}{\sum_j\exp(z_j/T)}\label{con:softmax}
\end{align}
where $N$ is the number of samples, $p$ is the prediction of student model, $z$ is the logit output of teacher model, $T$ is the temperature parameter controlling how much we rely on the BERT's prediction. 
We set $T=1$ in our experiment as suggested in \cite{jiao2019tinybert}.

\subsubsection{\textbf{Search Algorithm}}
We leverage the one-shot based architecture search algorithm~\cite{guo2019single} because it is one of the most effective and efficient methods among all state-of-the-art search algorithms. Weight sharing strategy is adopted to improve search efficiency in the one-shot model. Specifically, the search space $\mathcal{A}$ is encoded in a supernet, defined as $\mathcal{N}(\alpha, W)$, where $W$ are the weights of architecture. 
Any possible architecture $\alpha \in \mathcal{A} $ can be sampled from the supernet uniformly and inherit the same weight in their common graph nodes. Following~\cite{guo2019single}, those architectures consist of a series of blocks which have several choices of the operation. But only one choice is invoked in each block at the same time. 
One-shot approaches decouple supernet training and architecture searching in two sequential steps. In the first supernet training stage, the supernet is trained once following the teacher-student framework, which can be expressed as:
\begin{equation}
    W^* = \mathop{argmin}\limits_{W}\mathbb{E}_{\alpha \sim \Gamma(\mathcal{A})}\mathcal{L}_{KD}[\mathcal{N}(\alpha, W(\alpha)]
\end{equation}
\noindent where $\Gamma$ is a prior distribution and we set it as uniform distribution in our experiments. Thus, one architecture is sampled randomly in each step of optimization. 
Once the one-shot model has been trained, we use it to evaluate the performance of sampled structures.

The second stage is the architecture searching process. Architectures are picked uniformly and ranked by the knowledge distillation loss calculated on a split data. Importantly, a hard efficiency constraint is applied to filter the architectures with large memory occupation or high inference latency. The efficiency score of a given architecture could be computed by the following formula:
\begin{equation}
    \label{con:efficiency}
    \begin{split}
        & C \leq C_{max} \\
        & C=SIZE(\alpha)+TIME(\alpha)
    \end{split}
\end{equation}
where $SIZE(\cdot)$ and $TIME(\cdot)$ denote the normalized parameter size and inference time of architecture $\alpha$ respectively. $C$ is required to be no more than a preset budget $C_{max}$. If $C$ is larger than the threshold $C_{max}$ for a specific architecture, this candidate will not be considered. In our experiment, we set the value of $C_{max}$ according to the online efficiency constraint.

Before applying this architecture to production model, we use a large-scale real-world dataset collected from Microsoft Bing search log to retrain it for further improvement. The training details are described in the experiment section.

\begin{table*}[ht]
\small
  \renewcommand\arraystretch{1.2}
  \newcommand{\tabincell}[2]{\begin{tabular}{@{}#1@{}}#2\end{tabular}}  
  \begin{tabular}{c|c|c|c|c}
  \hline
  Query  & Ad Title & Ad URL & Ad Description & Label   \\ 
  \hline
  azure portal & Microsoft® Azure Portal  & azure.microsoft.com & Build, Manage, Monitor Everything from Simple to Complex & Good
  \\
  iphone & Microsoft PowerApps & powerapps.microsoft.com & Use Your Own Data to Create Sophisticated Apps  & Bad
  \\
  \hline
  \end{tabular}
  \caption{Ad Relevance Human Label Data Examples (content is simplified due to confidentiality)}
  \label{tab:labeldata-ads}
  \vspace{-0.8cm}
\end{table*}

\section{Experiments}

In this section, we describe our experiments on applying the proposed AutoADR framework to Ad Relevance. Details on datasets are provided first in the following sub-section. Then we conduct neural architecture search to find the best-performance achitecture. 
In section 5.3, we compare the derived model with baseline architectures, verifying its superiority in both performance and efficiency. Finally, we show offline and online results and analysis of integrating model learned by AutoADR to production Ad Relevance model.

\subsection{Teacher Data Generation}
There are two steps in generating teacher data used for neural architecture search and model retraining. The first one is to train a teacher model, the second one is to generate large-scale data scored by the teacher model. 

For teacher model training, we use Ad Relevance human label data to fine-tune BERT-large~\cite{devlin2019bert} model. The dataset contains millions of query-ad pairs 
labeled by professional human judges. Table \ref{tab:labeldata-ads} gives two examples of the label data. The ad with good label matches with query's intent, while the bad one does not.
Ad content consists of title, description, and URL. The label is based on relevance between query and overall comprehension of ad content. 
Therefore we use query string as input on query side, and concatenation of ad's title, description and normalized URL as input on ad side. 
We fine-tune uncased BERT large model with max sequence length 64, learning rate 1e-5, and training batch size 32. The model is fine-tuned for 2 epochs. 

We also collect a large-scale real-world dataset from Microsoft Bing search log, and use the fine-tuned BERT-large to do inference on those data to generate teacher scores. This dataset is shown as Train set in Table~\ref{tab:dataset-ads} which will be used for NAS search and retraining.
We also list the validation set for hyper-parameter search and test set for comparing AutoADR with baseline models in this table. Those two datasets are sampled from human label dataset.
\begin{table}[h]
\small
  \renewcommand\arraystretch{1.2}
  \newcommand{\tabincell}[2]{\begin{tabular}{@{}#1@{}}#2\end{tabular}}  
  \begin{tabular}{c|c|c|c}
  \hline
   Dataset  & Source & Volume & Supervision  \\ 
   \hline
  Train & Search log & 5m & Teacher score \\ 
  Validation & Label dataset & 200k & Human Label \\ 
  Test & Label dataset & 200k & Human Label \\
   \hline
  \end{tabular}
  \caption{Dataset for Neural Architecture Search and Retrain}
  \label{tab:dataset-ads}
  \vspace{-0.8cm}
\end{table}

\subsection{Neural Architecture Search}
In our experiments, we conduct neural architecture search on the Train set data to design a 6-layer optimal architecture. 
As mentioned in the table \ref{tab:dataset-ads}, the labels are the relevance scores predicted by the teacher model, thus as soft targets to guide the NAS process. For the search stage, we split a validation set consists of 200k samples from the Train data and the rest are used for the supernet training.

We train the one-shot model on the search space described in the previous section for about 2 days on 4 P100 GPUs. We set the batch size as 2,048, max query length as 16, max ad content length as 60, hidden unit dimension for each layer as 256, dropout ratio as 0.8 and L2 regularization as 2e-6. We utilize Adam optimizer for weights optimization. We adopt the cosine annealing learning rate decay, and the formula is:
\begin{equation}
    \vspace{-0.1cm}
    \lambda = \lambda_{min} + 0.5 \cdot (\lambda_{max} - \lambda_{min})(1 + cos(\pi T_{cur} / T))
\end{equation}
where $\lambda_{max}$ and $\lambda_{min}$ define the range of the learning rate, $T_{cur}$ is the current epoch number and $T$ is the cosine cycle. In our experiments, we set $\lambda_{max}=0.02$, $\lambda_{min}=0.00001$ and $T=10$. After training for 150 epochs, we randomly sample around 3,000 architectures from the search space and find the best architecture following the strategy described in section \ref{NAS_KD}.

\begin{figure}[t]
    \centering
    \includegraphics[width=\linewidth]{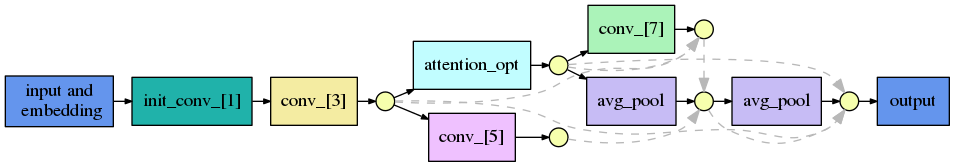}
    \caption{Visualization of the best architecture from AutoADR. Rectangles stand for layers, 
    one-way arrows stand for inputs and dotted arrows stand for shortcut connections.}
    \label{fig:arc}
    \vspace{-0.2cm}
\end{figure}

Figure \ref{fig:arc} visualizes the chosen architecture, which assembles multiple paths and different categories of layers, including 3 convolution layers, 2 avg-pooling layers, and 1 self-attention layer. The RNN layer is not included due to the latency constraint. CNN layers with small kernel size generate local information, which is located in early layer of searched network, while large-size kernel CNN layers are close to the output layer to capture long-term dependencies. The self-attention layer, as complementary to CNN layers, is capable to integrate global information. The design principles are in line with human common sense, which performs pooling and different convolution operations in parallel before aggregating them as final representation. 

\subsection{Result Comparison}

We retrain our architecture from scratch and compare it with several human designed models, including the convolution network (CNN), the recurrent network (RNN), Transformer (TRM), compact Transformer (Compact-TRM) and C-DSSM. 
Compact Transformer is a lighter Transformer, where the size of the feed-forward intermediate layer is set to be equal to the hidden unit dimension.
Among them, the structures of CNN, RNN and Compact-TRM are sub-networks of our search space, which have the opportunity of being selected during the search process. C-DSSM~\cite{shen2014latent} consisting of convolution and feed-forward layers, has proven effective in both retrieval and relevance tasks.

In our experiments, all the models are trained on the Train set, and critical hyper-parameters including batch size, learning rate and weight decay are decided according to the performance on the Validation set to make a fair comparison. Additionally, we evaluate different configurations of baseline models in \{64, 128, 256, 512\} for hidden unit dimension and \{2,4,6,8\} for the number of layers, and present the best result of each model in Table \ref{tab:comparasion_results}. 
Precision metric is PR AUC (Prediction-Recall Area Under Curve).
The results show that the AutoADR model outperforms all baseline methods, which demonstrates the superiority of neural architecture search. 
Notably, AutoADR with only 15.28 million parameters beats C-DSSM and compact Transformer by 4.47\% and 3.69\% respectively, while the improvement is 1.43\% compared with CNN. 

Moreover, we also compare the inference speed of all the models. Specifically, we evaluate the time to inference the whole Test set with the batch size as 128.
As shown in the table,  our model has comparable inference speed comparing to CNN, but it's much faster than non-CNN baselines. Specifically, AutoADR is $1.8\times$ and $2.2\times$ faster than C-DSSM and Transformer respectively. 
This result is within expectation. The network searched by the AutoADR framework is more efficient than most baseline models, as more convolutional layers and pooling layers are leveraged instead of recurrent and self-attention modules.

Based on the results in Table \ref{tab:comparasion_results}, we can conclude that the proposed AutoADR framework is capable of finding the network structure with less-complexity and better performance compared to human designed models. Next, we evaluate AutoADR in large-scale industrial scenarios, where the computation cost, especially online computation cost, is usually a bottle-neck for current state-of-the-art models.

\begin{table}[ht]
  \renewcommand\arraystretch{1.2}
  \begin{center}
  \begin{small}
  \begin{sc}
  \begin{tabular}{l|c|c|c}
    \hline
    Method & \#Params (million) & Inference Time (s) & AUC \\ 
    \hline
    CNN & 15.28 & 24.73 & 83.73 \\
    RNN & \textbf{7.25} & 31.09 & 81.73 \\
    TRM & 22.40 & 55.31 & 82.53 \\
    CompactTRM & 17.67 & 52.27 & 81.47 \\
    C-DSSM & 88.81 & 45.03 & 80.69 \\
    \hline
    AutoADR & 15.28 & \textbf{23.25} & \textbf{84.60} \\
    \hline
  \end{tabular}
  \end{sc}
  \end{small}
  \end{center}
  \caption{PR AUC Comparison}
  \label{tab:comparasion_results}
  \vspace{-0.5cm}
\end{table}

\subsection{Integration to Production Model}
In this section, we integrate AutoADR to our current production Ad Relevance model as a sub-model.
Before applying it to production, we re-train the best architecture learned by AutoADR with a larger data set with 500m query-ad pairs scored by the teacher model. After extending the amount of teacher data, it can retain 99.7\% of teacher's performance evaluated based on single feature PR AUC. This indicates AutoADR is efficient at capturing teacher knowledge within the knowledge distillation framework.

After that, we use the re-trained model to compute final AutoADR score and add it into current production model as a new input.
Table~\ref{tab:ads_full} shows results on our production evaluation sets after adding AutoADR. Here in this table, both TestSet-1 and TestSet-2 are human labeled query-ad pairs. TestSet-1 is sampled from online impressions, and TestSet-2 is sampled from Ad Retrieval component's output. Among these two, TestSet-1 is the major evaluation set for our production Ad Relevance model. Due to business confidentiality, numbers are shown as normalized PR AUC lift with respect to our shipping bar.

\begin{table}[ht]
\small
  \renewcommand\arraystretch{1.2}
  \newcommand{\tabincell}[2]{\begin{tabular}{@{}#1@{}}#2\end{tabular}}  
  \begin{tabular}{c|c|c}
  \hline
   Model & \tabincell{c}{TestSet-1} & \tabincell{c}{TestSet-2}   \\
  \hline
    Production + AutoADR & 2.65 & 5.01\\ \hline
  \end{tabular}
  \caption{Normalized PR AUC lift with respect to shipping bar after adding AutoADR into production model}
  \label{tab:ads_full}
  \vspace{-0.5cm}
\end{table}
As shown in the table, adding AutoADR can largely boost the production model's AUC performance. Since the current production model already contains many advanced sub-models and features, this is considered as a significant improvement that surpasses our shipping bar. 

In addition, we conduct thorough tests for hosting this AutoADR sub-model in production environments, and its memory and latency cost are well below our shipping constraints. This is expected since these constraints were already considered during AutoADR training process. Then we move on to the online A/B testing phase which will be described in next section.

\subsection{Online A/B Testing}
The online integration of AutoADR model into Ad Relevance model follows the mechanism mentioned in Section 3.2. 
We conduct online A/B testing for Ad Relevance models with and without AutoADR sub-model. 
The results are summarized in Table \ref{tab:online}. 
Here we show several key online metrics related to Ad Relevance (the numbers are normalized due to business confidentiality), including: 

\textbf{Bad-Ad ratio}: ratio of irrelevant ad impressions with respect to total ad impressions. In online flight, this ratio is approximated by sampling ad impressions and submitting them to human judges to get labels. This is our major online metric. 

\textbf{Click Yield}: average number of ad clicks per search result page view. Larger number indicates that ads shown to users can attract more clicks.  

\textbf{Quick Back Rate}: ratio of quick back ad clicks with respect to total ad clicks. A quick back ad click means user spends very short time on the ad's website page. It usually indicates that the user is not interested in the page's content and it’s good for search engine to reduce this rate.

Online A/B flights in Sponsored Search are complex and have many metrics. To evaluate Ad Relevance model improvements, we normally keep other metrics (like total ad impressions, overall revenue, etc.) at neutral levels and observe changes in those three aforementioned relevance-related metrics. In our online A/B flights, adding AutoADR results in 4.6\% Bad-Ad ratio reduction, which is statistically significant with a p-value of $0.00476$. This also surpasses our shipping bar to a large extent. In the meanwhile, we see improvements on Click Yield and Quick Back Rate metrics. Note that there are dedicated models to improve Click Yield and Quick Back Rate, Ad Relevance models can only impact these two metrics indirectly. Considering AutoADR training doesn't consider those two metrics, the improvements
confirm that ads on treatment flight are more relevant from user interaction perspective. In general, we can conclude that integrating AutoADR into production Ad Relevance model shows very positive impact on relevance-related online metrics.
We have shipped this technique in Microsoft Bing Ad Relevance Production model.
\begin{table}[ht]
\small
  \renewcommand\arraystretch{1.2}
  \newcommand{\tabincell}[2]{\begin{tabular}{@{}#1@{}}#2\end{tabular}}  
  \begin{tabular}{c|c|c}
  \hline
  Bad-Ad Ratio & Click Yield  & Quick Back Rate \\ 
  \hline
   \textbf{-4.60}\% & \textbf{+0.16}\%	& \textbf{-0.33}\% \\ 
  \hline
  \end{tabular}
  \caption{Online A/B testing result for Ad Relevance model with AutoADR}
  \label{tab:online}
  \vspace{-1cm}
\end{table}

\section{Conclusion}
In this paper, we propose AutoADR, a novel end-to-end framework for automatic model design with knowledge distillation, which encapsulates the privileges of AutoML and pre-training collaboratively. We conduct offline experiments to verify its outstanding effectiveness and efficiency compared to baseline models. In the online A/B testing phase, it shows a statistically significant 4.6\% Bad-Ad ratio reduction. This model has been shipped to the mainstream model of Microsoft Bing Ad Relevance. Moreover, AutoADR is a general framework that is not limited to the Ad Relevance scenario. As it has demonstrated its power for automatic model design for a specific task, we plan to apply the AutoADR framework to other production scenarios in the future. For example, search relevance, machine translation, and question answering. We will also keep exploring more advanced search algorithms to improve the effectiveness and efficiency of the neural architecture search procedure.

\bibliographystyle{ACM-Reference-Format}
\bibliography{CIKM2020_AutoADR_ref}

\end{document}